\documentclass[conference]{IEEEtran}
\IEEEoverridecommandlockouts
\usepackage{cite}
\usepackage{amsmath,amssymb,amsfonts}
\usepackage{algorithmic}
\usepackage{graphicx}
\usepackage{algorithm}
\usepackage{algorithm}
\usepackage{amsmath}
\usepackage{multirow}
\usepackage{textcomp}
\usepackage{xcolor}
\usepackage{svg}  %Sanaz added
\usepackage{todonotes}  %Sanaz added
\def\BibTeX{{\rm B\kern-.05em{\sc i\kern-.025em b}\kern-.08em
    T\kern-.1667em\lower.7ex\hbox{E}\kern-.125emX}}
\begin{document}

\title{Bit-Flip Fault Attack: Crushing Graph Neural Networks via Gradual Bit Search
% \thanks{Identify applicable funding agency here. If none, delete this.}
}
\author{\IEEEauthorblockN{Sanaz Kazemi Abharian}
\IEEEauthorblockA{\textit{Department of Electrical and Computer Engineering} \\
\textit{George Mason University}\\
Fairfax, VA, USA \\
skazemia@gmu.edu}
\and
\IEEEauthorblockN{Sai Manoj Pudukotai Dinakarrao }
\IEEEauthorblockA{\textit{Department of Electrical and Computer Engineering} \\
\textit{George Mason University}\\
Fairfax, VA, USA \\
spudukot@gmu.edu}
}
\maketitle

\begin{abstract}
Graph Neural Networks (GNNs) have emerged as a powerful machine learning method for graph-structured data. A plethora of hardware accelerators has been introduced to meet the performance demands of GNNs in real-world applications. However, security challenge of hardware-based attacks have been generally overlooked. In this paper, we investigate the vulnerability of GNN models to hardware-based fault attack, wherein an attacker attempts to misclassify output by modifying trained weight parameters through fault injection in a memory device. Thus, we propose Gradual Bit-Flip Fault Attack (GBFA), a layer-aware bit-flip fault attack, select a vulnerable bit in each selected weights gradually to compromise the GNN 's performance by flipping minimal number of bits. To achieve this, GBFA operates in two steps. First, a Markov model is created to predict the execution sequence of layers based on features extracted from memory access patterns, enabling the launch of the attack within a specific layer. Subsequently, GBFA identifies vulnerable bits within the selected weights using gradient ranking through an in-layer search. We evaluate the effectiveness of the proposed GBFA attack on various GNN models for node classification tasks using the Cora and PubMed datasets. Our findings show that GBFA significantly degrades prediction accuracy, and the variation in its impact across different layers highlights the importance of adopting a layer-aware attack strategy in GNNs. For example, GBFA degrades GraphSAGE’s prediction accuracy by 17\% on the Cora dataset with only a single bit flip in the last layer.
\end{abstract}

\section{Introduction} 

% \textcolor{red}{
% \begin{itemize}
%     \item Hardware aspect has to be improved, and also make sure to include EDA metrics in the results. 
%     \item Comparison with state-of-the-art (in results) 
%     \item Improve the state-of-the-art section (Recent papers and based on general BFAs - not limited to GCNs) 
%     \item Improvise Fig 1 
%     \item Assumption section 
%     \item Proofreading 
% \end{itemize}
% } 
Graph Neural Networks (GNNs) have emerged as a practical deep learning approach for graph-structured data, which is useful in various real-world applications such as social networks, electronic design automation (EDA) for accelerating hardware verification \cite{saravanan2024graphfuzz},\cite{gubbi2023hardware}, circuit security tasks \cite{chen2021deep,TCAD'22},\cite{Zhiqian_DATE'20}, and recommendation systems \cite{wu2022graph}. 
To address the limitations of software-based GNN 
implementations, hardware-based GNN accelerators are 
introduced in the recent years \cite{kose2024survey}. 
These GNN hardware accelerators have shown higher 
energy efficiency, lower latency, and improved performance 
compared to their software counterparts \cite{zhang2022low}.

GNNs involve specialized computations characterized by irregular and sparse graph structures, necessitating efficient hardware accelerators such as GPUs \cite{xie2023accel}, FPGAs \cite{tian2022fp}, and ASICs \cite{hu2022survey} to handle data movement, parallelize sparse computations, and minimize latency. Among these, custom ASIC accelerators offer significant advantages due to their tailored designs, which closely align with the irregular computational patterns of GNN workloads. Specifically, ASICs achieve lower latency, higher throughput, and substantially reduced energy consumption by optimizing data flow and integrating tightly coupled on-chip memory and processing logic. Unlike FPGAs, ASICs do not incur overhead from reconfigurable logic and routing, resulting in better performance density and improved power efficiency. Furthermore, ASIC accelerators often support 32-bit floating-point (FP32) arithmetic to preserve numerical precision during both training and inference phases. This design choice is reflected in several recent architectures, including HyGCN \cite{yan2020hygcn} and ReGNN \cite{chen2022regnn}, which adopt FP32 arithmetic to ensure stable and accurate computations.
\par Despite their superior performance and other benefits, 
recent works have shown that hardware accelerators 
%can be targeted% 
are vulnerable to hardware attacks \cite{mittal2021survey}, among which %targeted%
bit-flip attacks (BFAs) are considered 
one of the prominent attacks as they can crush a neural network through maliciously flipping a small number of bits within its weight parameters stored in memory. There are various methods, such as the Laser Beam attack and Row Hammer Attack (RHA), that perform bit-flip operations. RHA, introduced by Kim et al. \cite{kim2014flipping}, exploits a vulnerability in dynamic random access memory (DRAM), allowing an attacker to manipulate data without direct access. This is achieved by repeatedly "hammering" specific rows of memory cells, which induces bit flips in adjacent rows. While the
The impact of BFAs has been widely studied on Deep Neural Network (DNN) accelerators \cite{mittal2020survey}, their effect on GNNs remains largely unexplored, and the security risks posed by BFAs in GNN models have been overlooked. 
\par Software-based adversarial attacks on GNNs such as Evasion attacks and poisoning attacks %\textcolor{red}{XYZ}% 
\cite{jin2021adversarial} can be detected and mitigated using the existing techniques such as differential privacy and federated learning \cite{li2017understanding,Sanket_DATE'23}. However, 
the non-volatile nature of BFAs makes them hard to detect and analyze. 

%As GNNs and GNN accelerators are becoming more widely adopted, it is crucial to examine their security implications against BFA. 
%On the other hand, studies such as \cite{qian2023survey} have shown that the random injection of BFAs are inefficient, as the DNNs and GNNs exhibit inherent resilience against such random bitflips.  
%\textcolor{red}{Also, how is proposed attack different from RHA? Are there any targeted attacks on GNNs? If not, specify why having a targeted attack is crucial.}
%\textcolor{blue}{I modified these two paragraphs. Since we focus on hardware accelerators, I revised them to emphasize hardware attacks. Because our attack is untargeted, I removed the discussion on targeted attacks and did not include any explanation of them.}
\par In this work, we introduce a two-pronged approach for launching  %a targeted
 our proposed BFA on the GNN hardware accelerator.  First, we determine the GNN layer that is being executed by the hardware based on recorded memory patterns and use a Markov model to predict the layer sequence. 
Once the executing layer is identified, the proposed Gradual Bit-Flip Fault Attack (GBFA) on GNN models is introduced, which injects faults into trained weights using a layer-aware method that can cause a GNN to misclassify with a minimal number of bit flips. Our proposed GBFA attack gradually selects weights through an in-layer search based on the Bit Error Rate (BER) to ensure a decrease in prediction accuracy with a minimal number of bit flips.
\par We evaluate the effectiveness of the proposed GBFA Attack across multiple GNN architectures using two widely adopted benchmark datasets: Cora and PubMed \cite{kipf2016semi}. Our experimental results demonstrate that performing a targeted in-layer search is critical for maximizing the impact of GBFA while minimizing the number of bit flips. Additionally, our analysis reveals that the degree of accuracy degradation induced by GBFA varies across different GNN models, depending on the specific layer targeted. These findings highlight the critical impact of layer-wise vulnerability in enhancing the success of bit-flip fault attacks on GNNs. We observe that the attack consistently leads to a notable degradation in model accuracy across different GNN architectures. These results collectively validate the broad applicability and efficacy of GBFA as a stealthy and impactful fault injection method against GNNs.
%For instance, GBFA reduces GraphSAGE’s performance by 3.8\% with only 22 bits in layer 1, whereas it decreases performance by 16.46\% with only 1 bit in layer 2 on the Cora dataset. Additionally, its performance degrades by 2.6\% with 32 bits in layer 1, whereas it decreases by 7.79\% with only 1 bit in layer 2 on the PubMed dataset. 
% \textcolor{red}{Are you summarizing results here or what exactly is the message? It is confusing with lot of numbers. If you want to show impact on layer 1 and 2, provide average numbers - individual dataset numbers are tough to grasp in one reading.} 
% \textcolor{blue}{I modified the entire paragraph by removing the numbers and focusing on why we use in-layer attacks, as well as the use of different GNN models for evaluating the GBFA attack.}
%\textcolor{red}{Add a sentence on results!!}

The novel contributions of this work is summarized as follows: 

\begin{itemize} 
    \item We propose a hardware-based bit-flip fault attack on GNN models to degrade prediction accuracy on node classification task.
    \item We demonstrate that in-layer searches are essential for an effective bit-flip attack on GNNs. To identify and target a specific layer, our attack analyzes memory patterns and employs a Markov model, incorporating a Connectionist Temporal Classification (CTC) decoder to predict the run-time layer sequence.
    \item To achieve the desired accuracy degradation with a minimal number of bit flips, our proposed GBFA attack integrates gradient-based ranking and a gradual search within a specific layer to identify a vulnerable bit in each selected weight parameter. The bit-flip operations are then performed along the gradient 's ascending direction.
%     \item We evaluate GBFA on a wide range of GNN models, including GCN, GAT, GraphSAGE, and GIN, for the node classification task using the Cora and PubMed datasets. Additionally, we compare our proposed method with a random bit-flip fault attack. Our results demonstrate the effectiveness of our attack in compromising model performance.
\end{itemize} 
\begin{figure} [!h]
     \centering
     \includegraphics[width= \linewidth]{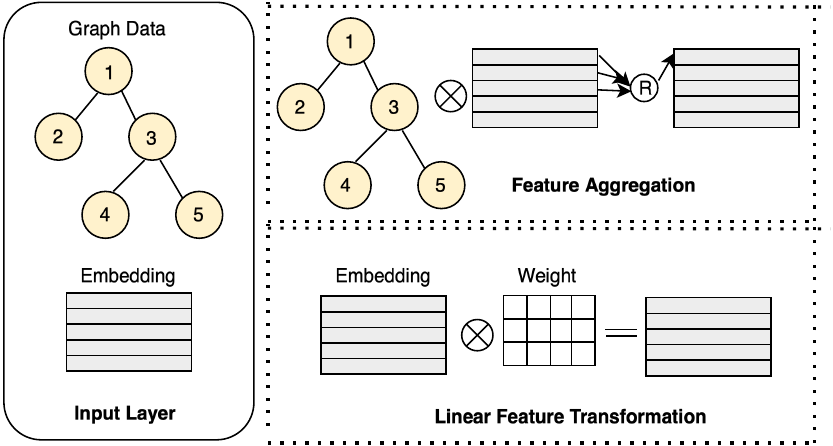}
     \caption{Computational procedure of a single GCNConv layer} 
      \label{1}
\end{figure}
\section{Background and Related Works}
\subsection{Graph Neural Networks}
GNNs are a class of neural networks designed to operate on graph-structured data \cite{rafatirad2022graph}. Graph Convolutional Networks (GCNs), constructed using GCNConv layer, operate in two fundamental phases feature aggregation and linear feature transformation as shown in Figure \ref{1}. The forward pass in the l-th GCNConv layer is functioned into two stages: 1) a feature aggregation step followed by 2) feature transformation step. Each node aggregates feature information from its neighbors through an aggregation function, capturing local structural dependencies. The resulting aggregated feature vectors are then passed through a combination function, typically parameterized by shared trainable weights to produce updated node representations \cite{etemadyrad2024global}.
\par Variants of GCNs, such as GraphSAGE, Graph Isomorphism Network (GIN), Graph Attention Network (GAT), retain the general message-passing framework but differ in their neighborhood aggregation functions, while still conforming to the fundamental propagation scheme of standard GCNs.
%\textcolor{red}{Use the terms used in Fig 1 and also refer to that. For instance, the term feature vector is not discussed in the explanation, but is in the figure. Same with combination. You have update phase, but is not indicated in Fig 1. }
%\textcolor{blue}{I revised the paragraph according to your comment.}
\subsection{Related Works}
%In this section, we briefly discuss some of the recent works on bit-flip attacks on hardware accelerators (DNN and GNNs). 
%Bit-flip faults in memory devices pose a significant threat to computing systems. Recent advancements in memory fault injection techniques have enabled adversarial weight attacks on neural networks deployed on computing platforms. 
\par Recent research demonstrates the vulnerability of Deep neural network (DNN) parameters stored in DRAM to bit-flip fault attacks. Liu et al. \cite{liu2017fault} proposed a Single Bias Attack (SBA), which targets a specific bias term of a neuron to cause misclassification in a DNN. The results demonstrate that even modifying a single bias can result in misclassification. Hong et al. \cite{hong2019terminal} investigate the impact of bit-flip fault attacks on the weights of various DNN architectures with full precision, leveraging the Row Hammer method. Their study examines how bit position, flip direction, parameter sign, and layer width influence the effects of such attacks. In \cite{rakin2019bit}, Rakin et al. introduced the progressive bit search (PBS) method, which leverages the Rowhammer technique to target the quantized weight parameters of DNNs. PBS ranks and progressively searches for the most vulnerable bits, flipping them in a way that maximizes accuracy degradation. The effectiveness of this attack is investigated on ResNet-18 using the ImageNet dataset, causing a significant drop in accuracy.
\par In \cite{kummer2024attacking}, the injective bit flip attack (IBFA) was introduced on a quantized GNN model. The IBFA targets bit flipping in the injective properties of the neighborhood aggregation function in message passing layers to reduce their ability to distinguish non-isomorphic graphs and to impair the expressivity of the Weisfeiler-Leman test. 
%In addition, Kummer et.al in \cite{kummer2025crossfire} propose a defense method, Crossfire, against BFAs on GNNs. The Crossfire employs the combination of hashing, honeypots, and bit-level correction for defense mechanism. This paper uses the Honeypot method to select important weights that attract bit-flip attacks. However, the paper only applied the model on GIN.
\par \textbf{Limitation of previous works}. Prior research on BFAs has primarily focused on DNNs, with limited attention to their impact on GNNs. The unique computational structure of GNNs warrants further investigation to assess their vulnerability to BFAs. Additionally, while studies \cite{malekzadeh2021impact}–\cite{wang2022pygfi} demonstrate that bit-flip faults at different layers affect the output differently, most fault attacks overlook layer-specific effects. To address this, our proposed GBFA attack focuses on in-layer search. 
\section{Proposed Hardware-based Bit-Flip Fault Attack on ML Accelerators}
\par The proposed GBFA is a gray-box fault-injection attack in which the adversary has physical access to the hardware accelerator executing the target GNN model. Furthermore, the adversary has no knowledge of the GNN model, training dataset, or hyperparameters. However, it is assumed that the adversary has access to the gradients and the test dataset. For each trained weight subjected to the fault attack, only a single bit can be affected. Additionally, it is assumed that the training process and its parameters are fault-free prior to the attack. 
\subsection{Design of GNN Accelerator} 

As aforementioned, a plethora of hardware accelerators for GNNs are proposed in the 
literature \cite{zhang2023survey}. In this work, we consider Reconfigurable GNN accelerator \cite{chen2022regnn} as our base design and adapt it to the proposed GNN models and datasets. 

\begin{figure} [t]
     \centering
     \includegraphics[width= \linewidth]{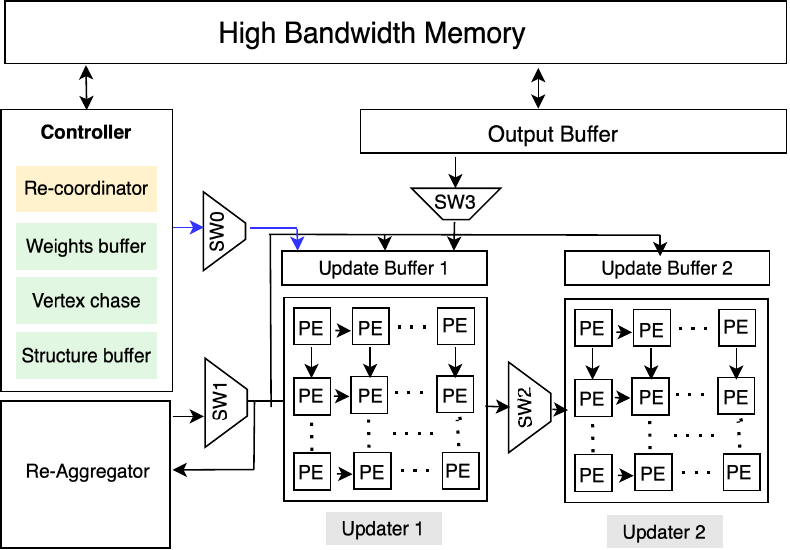}
     \caption{The overview of the Hardware accelerator}
      \label{2}
\end{figure}

The GNN accelerator shown in Figure \ref{2} is a configurable, pipelined hardware accelerator, designed to efficiently execute GNNs by eliminating computational and communication redundancies. The architecture contains three primary processing modules: Re-coordinator, Re-aggregator, and Flexible Updaters containing processing elements (PE), which are coordinated by a central controller and interconnected via a dynamic switch network. 
\begin{figure*}[t]
     \centering
      \includegraphics[width=0.8\linewidth]{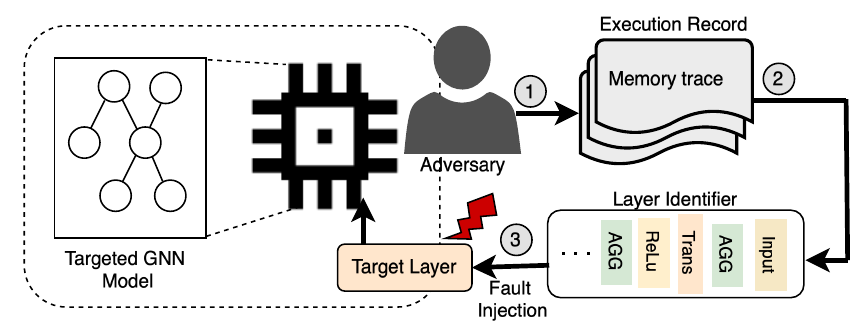}
      \caption{The overview of the proposed GBFA Attack}
       \label{3}
\end{figure*}
The Re-coordinator is responsible for scheduling tasks based on a redundancy-aware graph structure, identifying opportunities to reuse intermediate computations. It consists of a structure prefetcher, vertex prefetcher, and a dispatcher that dynamically allocate EdgeUpdate and Aggregation workloads while minimizing redundant data movement. The Re-aggregator performs feature aggregation for both target vertices and redundant neighbor sets using a parallel aggregation engine supported by a neighbor shuffler and task buffer; it caches intermediate aggregation results to reduce computation and communication redundancy. The Flexible Updaters implement matrix-vector multiplications for both EdgeUpdate and VertexUpdate using systolic arrays. In GNNs without EdgeUpdate, these arrays can be reconfigured and combined to maximize utilization. A central controller interprets a three-bit configuration word, indicating the presence of EdgeUpdate, EdgeUpdate Redundancy (ER), and Aggregation Redundancy (AR), to reconfigure  micro-configurable switches (SW0-SW3) and dynamically adapt the execution pipeline to various GNN variants. %\textcolor{red}{Point all these terms in Figure too.}

\par To support high-throughput and energy-efficient execution, ReGNN incorporates a hierarchical memory system comprising both on-chip and off-chip components. Off-chip memory is implemented using high-bandwidth memory (HBM) with a bandwidth of 256 GB/s to store vertex features, edge information, and weights. The on-chip memory includes specialized buffers for EdgeUpdate, Aggregation, weights, and structural metadata, as well as a configurable vertex cache. This cache stores original vertex features when redundancy elimination is not applied, and caches intermediate EdgeUpdate and Aggregation results when ER and AR are present, respectively.

To improve cache efficiency, ReGNN employs a priority-based management strategy, giving preference to high-degree vertices and larger redundant sets. Additionally, double buffering is used across memory modules to hide data transfer latency and overlap communication with computation. This memory architecture ensures minimal DRAM access, maximized data reuse, and improved pipeline utilization, contributing significantly to ReGNN’s performance and energy efficiency.
\subsection{Deploying GBFA on Hardware Accelerator} 
Figure \ref{3} illustrates the overall workflow of the proposed GBFA attack, which comprises two primary stages: identifying the target layer and injecting bit-flip faults into the selected weight parameters via an in-layer search within the targeted layer.
\subsubsection{\textbf{Run-time Layer Sequence Prediction}}
Input layer, message passing layer, transformation layer, readout layer, and output layer are some of the common layers in GNN models \cite{liu2020towards}.
Each layer of a GNN model exhibits a distinct memory access pattern and computational cycles per kernel at runtime. These characteristics make the model vulnerable to hardware snooping attacks, enabling an attacker to observe execution patterns. 
By leveraging this information, the adversary can predict and infer individual layers without access to the neural network’s parameters. 
\par \textbf{Record Memory Patterns:} An attacker can access GDDR bus to extract memory patterns, either by physically probing the interconnect or utilizing a DMA-capable device to achieve kernel events. The GDDR bus can reveal kernel events and memory copy size $(Mem_{cp})$, which can be further leveraged to infer execution latency $(Exe_{lat})$. Similarly, monitoring the device memory bus allows the extraction of memory request traces, from which key memory-related parameters, such as the number of read and write operations $(R_v, W_v)$ and raw data dependencies $d_{RAW}$  can be inferred. This architectural information leakage is used to form feature inputs for the HMM model.

Architectural features of kernels, such as kernel execution time ($Exe_{lat}$), kernel read volume ($R_v$), write volume through the memory bus ($W_v$), Input/output data volume ratio of each kernel ($I_v/O_v$), where the output volume ($O_v$) is measured by the write volume of this kernel, and input volume ($I_v$) is measured by the write volume of the previously executed kernel, kernel dependency distance ($kdd$) represents the topology influence. kdd is defined as the maximum distance between a given kernel and its preceding dependent kernels during the execution of a kernel sequence. This metric encodes layer structure information within the kernel features which can be computed as $kdd \approx \max(d_{\text{RAW}})$. These features are transformed into kernel features by extracting tracking memory patterns.
\par \textbf{A Layer Sequence Predictor:} Once the memory access patterns are recorded, we introduce a layer sequence predictor for predicting the GNN's layer. 
To build a model to predict layers, two primary information sources are considered: architectural kernel execution features and the distribution of layer context within the sequence. Thus, the runtime-layer sequence prediction problem can be formalized as a sequence-to-sequence model prediction which is described as follows: 

The input is kernel execution feature sequence $\vec{X}_t$ with temporal length of $T$ described as an array:
$(Exe_{lat}, R_v, W_v, I_v/W_v, \text{kdd})_t$ obtained from the memory access pattern observation. 
As the proposed GNN layer prediction involves analyzing temporal memory access patterns, we devise 
% The goal is to create 
a Hidden Markov Model (HMM) incorporating a Connectionist Temporal Classification (CTC) decoder. 
% to predict layer sequence. 
The Markov Model provides a probabilistic method where each state represents a specific layer, and transitions between states capture the probability of execution order based on the observed hardware characteristics i.e., memory access patterns. 
Thus, we formulate the object function of HMM layer sequence identifier is to minimize the CTC cost for a given target layer sequence $L^*$ which is computed as:
\begin{equation}
    cost(X) = -\log P(L^*|h(X))
\end{equation}
The $P(L^*|h(X))$ represents the probability of observing result $L^*$ given the input $X$. 

%$\mathbf{X} = (X_1, \dots, X_T)$,
%At each time step, the output vector is transformed to a probability distribution of the next layer OPs.

As a case in point, there is a sequence consisting three execution kernels. At each time step $(t_0, t_1, and t_2)$, the HMM produces a probability distribution over the layer operations and CTC decoder employs beam search to determine the sequence with the highest probability. 
\subsubsection{\textbf{Layer-aware Fault Attack Injection}} 
GBFA detects a vulnerable bit in a weight using a method similar to FGSM \cite{goodfellow2014explaining}. It utilizes the gradient ascent direction w.r.t. the loss function to rank bits. 
Thus, it calculates the gradient of a weight b, represented in binary format, according to Equation \ref{eq1}:
\begin{equation}
\nabla_{\mathbf{b}} \mathcal{L} = 
\left[ \frac{\partial \mathcal{L}}{\partial b_{31}}, \dots, \frac{\partial \mathcal{L}}{\partial b_0} \right]
\label{eq1}
\end{equation}
Where $\mathcal{L}$ represents the inference loss function of GNN parameterized by $b$. 
To flip the bit, the straightforward approach involves directly flipping bits based on the gradients derived in Equation \ref{eq2}, resulting in the perturbed bits:
\begin{equation}
\hat{\boldsymbol{b}} = \boldsymbol{b} + \operatorname{sign}(\nabla_{\boldsymbol{b}} \mathcal{L})
\label{eq2}
\end{equation}
However, this may lead incorrect result because of overflow.
%Applying the gradient directly to flip the bit, as described in equation \ref{eq2}, may lead to overflow and render the attack ineffective. Therefore, the attacker determines whether to flip the bit based on Table I, where I is an indicator specifying whether the bit-flip operation should be performed. Then, flip the bit according Equation \ref{eq3}.
GBFA decides to flip the bit based on Table \ref{T1} in order to increase the loss function. $I$ is an indicator that specifies whether the bit-flip operation should be performed. 
Thus, this is formulated in Equation \ref{eq3}.
%\begin{equation}
 %\hat{\mathbf{b}} = \mathbf{b} + \operatorname{sign}%(\nabla_{\mathbf{b}} \mathcal{L})
 %\label{eq2}
%\end{equation}
\begin{equation}
   \hat{b}_i = \operatorname{sign} \left( \frac{\partial \mathcal{L}}{\partial b_i} \right) \wedge (b_i \vee I) 
   \label{eq3}
\end{equation}
\begin{table}[h]
    \centering
    \caption{GBFA Truth Table depicting whether the value of $b_i$ alters or not}
    \renewcommand{\arraystretch}{1.3}
    \begin{tabular}{|c|c|c|c|}
        \hline
        $b_i$ & $\operatorname{sgn}(\partial \mathcal{L} / \partial b_i)$ & $\hat{b}_i$ & $I$ \\
        \hline
        0 & $+1$ & 1 & 1 \\
        0 & $-1$ & 0 & 0 \\
        1 & $+1$ & 1 & 0 \\
        1 & $-1$ & 0 & 1 \\
        \hline
    \end{tabular}
    \label{T1}
\end{table}

\begin{algorithm}
\floatname{algorithm}{\textbf{Algorithm}}
\small
%\footnotesize 
\caption{Layer-aware Fault Attack Injection}
    \label{al}
    \renewcommand{\algorithmicrequire}{\textbf{BitFlip} (GNN)}
      \begin{algorithmic}[1]
        \REQUIRE 
        \STATE Target a Layer
         \STATE Set BER
        \FOR{$i$ in range BER}
            \STATE Select weights randomly in the target layer
        \ENDFOR
        \WHILE {($\text{num\_weights}  != \text{BER}$)} 
            \STATE Compute $\frac{\partial L}{\partial w} = [\frac{\partial L}{\partial b_n}...\frac{\partial L}{\partial b_0}]$
            \IF{$(b_i = 0 \land \text{sgn}(\frac{\partial L}{\partial w}) > 0) \lor (b_i = 1 \land \text{sgn}(\frac{\partial L}{\partial w}) < 0)$}
                \STATE Flip the bit
            \ELSE
                \STATE Ignore the weight
            \ENDIF
        \ENDWHILE
        \STATE Conduct a performance evaluation of the GNN model with modified weights.
        \IF {attack is successful}
            \STATE End procedure
        \ELSE
            \STATE set another BER or target another layer
        \ENDIF
   \end{algorithmic}
\end{algorithm}

\par After determining the executing layer of the GNN in the previous step, GBFA identifies a vulnerable bit in the selected weight through an in-layer search, as described in Algorithm 1. In each iteration, the gradient of bits in the selected weight is computed and ranked. Then, the bit with the highest gradient and the corresponding I value, according to Table \ref{T1}, is selected for flipping. This targeted bit-flip operation is followed by an inference phase, where GBFA evaluates the loss function and the network's test accuracy after perturbation. The resulting accuracy measurements are recorded to generate a test accuracy profile, which reflects the progressive degradation of inference performance. It gradually selects weights based on the BER and evaluates the degradation of inference accuracy to ensure a significant decline while maintaining minimal BER. 
\begin{table}[htbp]
\caption{Graph Dataset Characteristics}
\begin{center}
\begin{tabular}{|c|c|c|c|c|}
\hline
Dataset& Nodes&  Edges& Feature Dimentions&  Categories\\
\hline

Cora& 2708& 5429& 1433&7 \\

PubMed& 19717& 44338& 500&3 \\
\hline
\end{tabular}
\end{center}
 \label{T2}
\end{table}

%\begin{table}[htbp]
%\caption{}
%\centering
%\resizebox{\columnwidth}{!}{%
%\begin{tabular}{|c|c|c|ccc|}
%\hline
%Dataset & GNN Models & Baseline Test Accuracy & \multicolumn{3}{c|}{Number of Trained Weights} \\
%\cline{4-6}
 %&  &  & Layer 1 & Layer 2 & Layer 3 \\
%\hline
%\multirow{4}{*}{Cora} & GCN & 0.81 & 91712 & 2048 & 224 \\
% & GAT & 0.78 & 91904 & 4288 & 469 \\
% & GraphSAGE & 0.798 & 22928 & 112 & - \\
% & GIN & 0.64 & - & - & - \\
%\hline
%\multirow{4}{*}{PubMed} & GCN & 0.7830 & 64000 & 8192 & 192 \\
% & GAT & 0.7740 & 128000 & 65536 & 768 \\
% & GraphSAGE & 0.769 & 32000 & 192 & - \\
% & GIN & 0.613 & - & - & - \\
%\hline 
%\end{tabular}%
%} % End of resizebox
%\end{table}
\begin{table}[htbp]
\caption{Accuracy and Model Size of GNN Models}
\centering
\resizebox{\columnwidth}{!}{%
\begin{tabular}{|c|c|c|c|}
\hline
Dataset & GNN Models & Baseline Test Accuracy & Total Trained Weights \\
\hline
\multirow{4}{*}{Cora} 
 & GCN & 0.810 & 93984 \\
 & GAT & 0.780 & 96661 \\
 & GraphSAGE & 0.798 & 23040 \\
 & GIN & 0.640 & 108070 \\
\hline
\multirow{4}{*}{PubMed} 
 & GCN & 0.783 & 72384 \\
 & GAT & 0.774 & 194304 \\
 & GraphSAGE & 0.769 & 32192 \\
 & GIN & 0.613 & 48360 \\
\hline 
\end{tabular}%
} % End of resizebox
\label{T3}
\end{table}

%\begin{table}[h]
 %   \centering
 %   \caption{Prediction Error Rate on GNN Models}
  %  \begin{tabular}{|c|c|c|c|c|c|c|c|}
  %      \hline
   %     GCN-Cora & GCN-PubMed & GAT-Cora & GAT-PubMed & GraphSAGE-Cora & GraphSAGE-PubMed & GIN-Cora & GIN-PubMed \\
    %    \hline
   %     0.032 & 0.050 & 0.020 & 0.023 & 0.060 & 0.062 & 0.043 & 0.045 \\
    %    \hline
 %   \end{tabular}
%    \label{T4}
%\end{table}
\begin{table}[h]
    \centering
    \caption{Prediction Error Rate on GNN Models}
    \begin{tabular}{|c|c|}
        \hline
        \textbf{Model} & \textbf{Error Rate} \\
        \hline
        GCN-Cora & 0.032 \\
        GCN-PubMed & 0.050 \\
        GAT-Cora & 0.020 \\
        GAT-PubMed & 0.023 \\
        GraphSAGE-Cora & 0.060 \\
        GraphSAGE-PubMed & 0.062 \\
        GIN-Cora & 0.043 \\
        GIN-PubMed & 0.045 \\
        \hline
    \end{tabular}
    \label{T4}
\end{table}

\begin{table*}[htbp]
    \begin{center}
        \caption{Evaluation of GBFA performance on GNN models using Cora and PubMed datasets at BER = 1e-1}
        \begin{tabular}{|c|c|c|c|c|c|c|c|c|c|}
            \hline
            \multirow{2}{*}{\textbf{GNN Model–Dataset}} & \multicolumn{3}{c|}{\textbf{Layer 1}} & \multicolumn{3}{c|}{\textbf{Layer 2}} & \multicolumn{3}{c|}{\textbf{Layer 3}} \\
            \cline{2-10}
            & PAC & ASR & $n_{bit}$ & PAC & ASR & $n_{bit}$ & PAC & ASR & $n_{bit}$ \\
            \hline
            GCN–Cora & 68\% & 23\% & 917 & 24\% & 70\% & 204 & 21\% & 73\% & 22 \\
            \hline
            GCN–PubMed & 41\% & 63\% & 6400 & 41\% & 53\% & 819 & 40\% & 42\% & 19 \\
            \hline
            GAT–Cora & 37\% & 57\% & 9190 & 40\% & 58\% & 428 & 22\% & 76\% & 46 \\
            \hline
            GAT–PubMed & 48\% & 46\% & 12800 & 47\% & 48\% & 6 & 63\% & 27\% & 7 \\
            \hline
        \end{tabular}
        \label{T5}
    \end{center}
\end{table*}
\begin{table*}[t]
\centering
\caption{Evaluation of GBFA Attack on GraphSAGE model at the lowest BER and at BER of 1E-1 
}
\begin{tabular}{|c|c|c|c|c|c|c|c|}
\hline
\textbf{GNN Model- Dataset} & \textbf{Layer} & \textbf{PAC (Min BER)} & \textbf{Min BER} & $\mathbf{n_{bit}}$ (Min BER) & \textbf{PAC (1e-1)} & \textbf{ASR (1e-1)} & $\mathbf{n_{bit}}$ (1e-1) \\
\hline
\multirow{2}{*}{GraphSAGE-Cora} 
& 1 & 76\% & $1\text{e}{-3}$ & 22 & 35\% & 62\% & 22928 \\
& 2 & 66\% & $1\text{e}{-2}$ & 1  & 13\% & 88\% & 22 \\
\hline
\multirow{2}{*}{GraphSAGE-PubMed} 
& 1 & 75\% & $1\text{e}{-3}$ & 32 &42\% & 56\% & 640 \\
& 2 & 71\% & $1\text{e}{-2}$ & 1  & 40\% & 58\% & 38 \\
\hline
\end{tabular}
\label{T6}
\end{table*}
\begin{table*}[t]
\centering
\caption{Evaluation of GBFA Attack on GIN model at the lowest BER and at BER of 1E-1 
}
\begin{tabular}{|c|c|c|c|c|c|c|c|}
\hline
\textbf{GNN Model- Dataset} & \textbf{Layer} & \textbf{PAC (Min BER)} & \textbf{Min BER} & $\mathbf{n_{bit}}$ (Min BER) & \textbf{PAC (1e-1)} & \textbf{ASR (1e-1)} & $\mathbf{n_{bit}}$ (1e-1) \\
\hline
\multirow{5}{*}{GIN-Cora} 
& 1 & 58\% & $1\text{e}{-2}$ & 917 & 12\% & 85\% & 9171 \\
& 2 & 55\% & $1\text{e}{-2}$ & 40  & 13\% & 83\% & 409 \\
& 3 & 33\% & $1\text{e}{-2}$ & 40  & 13\% & 85\% & 409 \\
& 4 & 52\% & $1\text{e}{-2}$ & 40  & 15\% & 83\% & 409 \\
& 5 & 60\% & $1\text{e}{-1}$ & 40  & 60\% & 5\%  & 409 \\
\hline
\multirow{5}{*}{GIN-PubMed} 
& 1 & 60\% & $1\text{e}{-2}$ & 320 & 57\% & 22\% & 3200 \\
& 2 & 59\% & $1\text{e}{-2}$ & 40  & 56\% & 22\% & 409 \\
& 3 & 60\% & $1\text{e}{-2}$ & 40  & 56\% & 19\% & 409 \\
& 4 & 60\% & $1\text{e}{-2}$ & 40  & 58\% & 12\% & 409 \\
& 5 & 60\% & $1\text{e}{-2}$ & 40  & 58\% & 8\%  & 409 \\
\hline
\end{tabular}
\label{T7}
\end{table*}

%\subsection{Detection of ML Layer Execution} 
%\subsection{Layer-aware Fault Attack Injection} 
%\begin{figure} [!h]
  %  \centering
  %   \includesvg[width= \linewidth]{GCNGAT.svg}
  %  \caption{Average post-attack test accuracy of the GCN and GAT models on the Cora and PubMed datasets under the GBFA attack at the minimum BER that results in a performance drop. \textcolor{red}{Font size}}
   %   \label{2}
%\end{figure}
\begin{figure} [t]
     \centering
     \includegraphics[width= \linewidth]{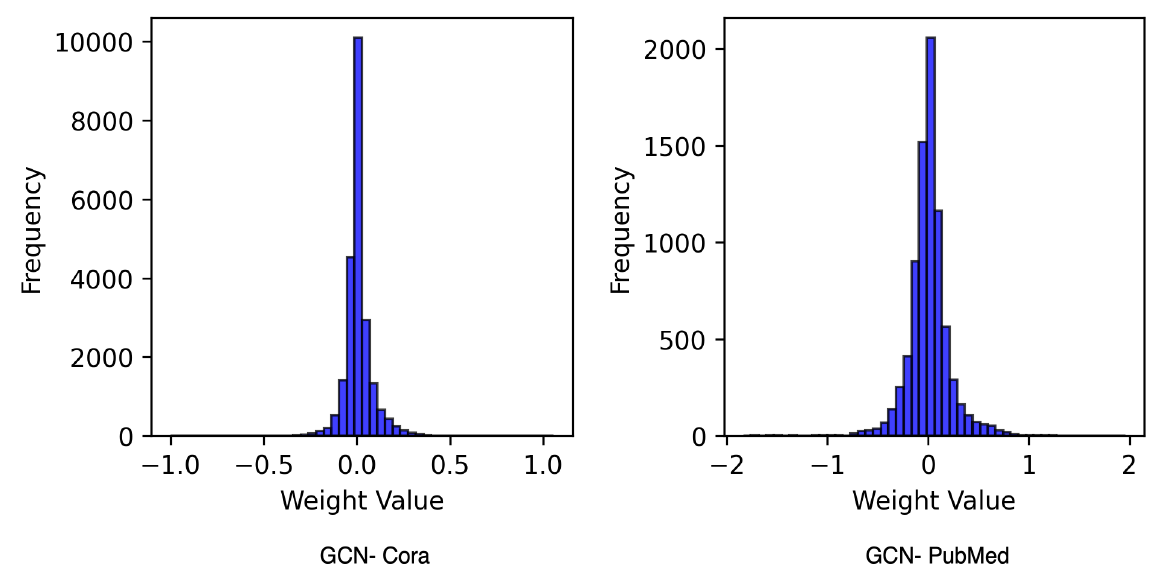}
     \caption{Weight distribution of GCN trained on the Cora and PubMed datasets}
      \label{4}
\end{figure}
\begin{figure} [t]
     \centering
     \includegraphics[width= \linewidth]{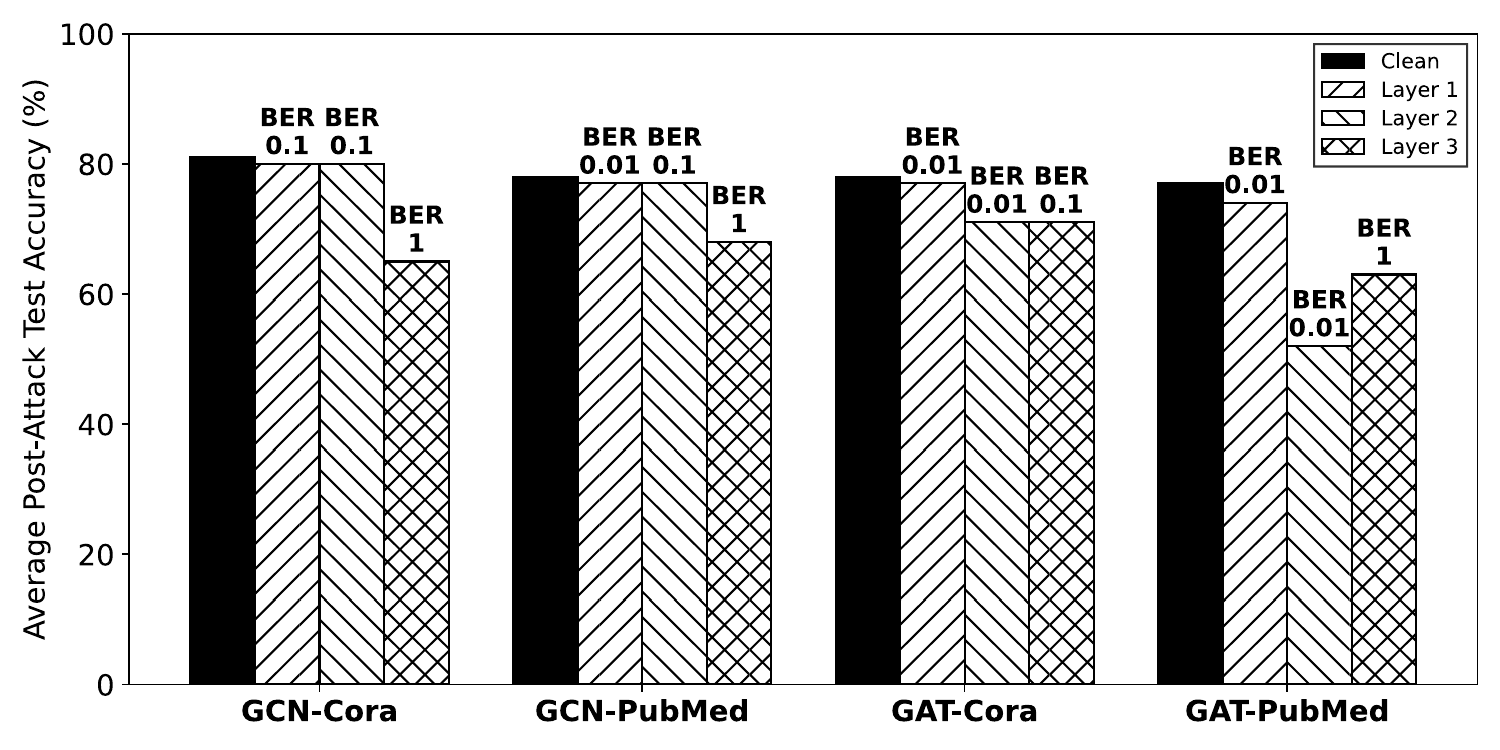}
     \caption{Average post-attack test accuracy of the GCN and GAT models on the Cora and PubMed datasets under the GBFA attack at the minimum BER that results in a performance drop.}
      \label{5}
\end{figure}
\section{Results and Discussion}
\subsection{Experimental Setup}
\textbf{Datasets.} We evaluate the effectiveness of the GBFA attack on node classification tasks using two benchmark datasets, as summarized in Table \ref{T2}. These datasets are chosen for their diversity in size and structural complexity: Cora is a small-scale dataset, whereas PubMed is a large one. Additionally, Cora has high-dimensional sparse features, while PubMed is denser. Moreover, since node classification is a semi-supervised learning task, the label rate is 0.052 for the Cora dataset, and 0.003 for the PubMed dataset, meaning that only about 5.2\% of Cora, and 0.3\% of PubMed data is labeled for training. 
\par \textbf{GNN Models.} We evaluate the effectiveness of the GBFA attack on four widely used GNN models: GCN, GAT, GraphSAGE, and GIN models. We employed PyTorch \cite{paszke2019pytorch} and the PyTorch Geometric (PyG) \cite{fey2019fast} libraries to train GNNs and evalauate GBFA attack. In Table \ref{T3}, we summarize the test accuracy of the fault-free models and provide the number of trained weights in each layer of the GNN models. Moreover, Figure \ref{4} illustrates the weight distribution of the GCN model on the Cora and PubMed datasets, which is representative of the weight range observed in other GNN models. The Cora dataset results in a more constrained weight range compared to PubMed.
\par \textbf{Evaluation Metrics:}
We evaluate the performance of the GBFA attack using three metrics: post-attack test accuracy (PAC), the number of flipped bits $(n_{bit})$, and the Attack Success Rate (ASR). Post-attack test accuracy refers to the percentage of test data correctly classified by the GNN model after the attack. $n_{bit}$ represents the number of bits flipped by the attacker to reduce the model’s accuracy at the defined BER. ASR quantifies the proportion of test samples whose predicted labels are altered as a result of an induced bit-flip fault.
%top-k accuracy
\subsection{Layer Sequence Prediction Performance}
In this section, we assess the accuracy of layer sequence prediction using the layer prediction error rate (LER) defined in Equation \ref{5} \cite{hu2020deepsniffer}. LER is computed as the mean normalized edit distance between the predicted and actual sequence. 
\begin{equation}
LER = \frac{ED(L, L^*)}{|L^*|}
 \label{5}
\end{equation}
\par $ED(L, L^*)$ represents the edit distance between the predicted layer sequence $L$ and the ground-truth layer sequence $L^*$. It is defined as the minimum number of insertions, deletions, and substitutions needed to transform $L$ to $L^*$. The term $|L^*|$ denotes the length of the ground-truth layer sequence. Table \ref{T4} represents the LER of GNN models.
\subsection{Evaluating the effectiveness of GBFA on various GNN models}
Figure \ref{5} illustrates the average post-attack test accuracy of GCN and GAT models on Cora and PubMed datasets under the GBFA at minimum BER that result in a performance drop. The results indicate that the BER value varies across models and layers. However, deeper layers in both models are more vulnerable to GBFA attacks.
 For example, GBFA can achieve 20\% accuracy drop by injecting faults in layer three with minimum BER.
The summary of evaluation of GBFA at BER of 1e-1 is presented in table \ref{T5}. The results indicate that deeper layers are more susceptible to the attack. Moreover, at a BER of 1e-1, GBFA  achieves an average ASR of 75\% on GCN and GAT using the Cora dataset by focusing on layer three.
\par Table \ref{T6} presents the effectiveness of the GBFA attack on the GraphSAGE model using the Cora and PubMed datasets. For both datasets, flipping only one vulnerable bit in layer 2 is sufficient to degrade the accuracy. Moreover, GBFA achieves an ASR of 88\% on the Cora dataset and 58\% on the PubMed dataset by targeting layer 2 at a BER of 1e-1.
\par Table \ref{T7} presents a comprehensive evaluation of the GBFA attack on the GIN model across individual layers for the Cora and PubMed datasets, under both the minimum BER and BER of 1e-1. For GIN-Cora, the attack demonstrates varying degrees of performance accuracy collapse (PAC) depending on the targeted layer. Notably, Layer 3 exhibits the most significant drop in performance at Min BER, with PAC falling to 33\%, while Layer 5 maintains a relatively high PAC of 60\% at a higher BER of 1e-1, accompanied by a notably low attack success rate (ASR) of only 5\%. This suggests layer-specific resilience differences. In contrast, for GIN-PubMed, the PAC values remain stable at Min BER across all layers, while under a BER of 1e-1, PAC declines moderately (56–58\%) with ASR values ranging from 22\% in the earlier layers to 8\% in the final layer. These results indicate that the GBFA’s impact is highly dependent on both the targeted layer and dataset, with deeper layers in GIN-Cora being more vulnerable and consistent PAC-ASR trade-offs observed in GIN-PubMed.
\par \textbf{GBFA Effectiveness Across GNN models:} 
\par The susceptibility of GNN models to the GBFA attack varies significantly based on their architectural design and depth. We observe that model-specific factors such as layer composition, aggregation mechanisms, and parameter count influence the severity of accuracy degradation under bit-flip faults.
\par Among the evaluated architectures, GIN and GraphSAGE consistently exhibit higher vulnerability. In particular, GIN suffers sharp accuracy drops even at low BER values. This heightened sensitivity stems from GIN’s reliance on precise weight updates to maintain injective aggregation, making it particularly fragile to targeted bit-level perturbations.
\par GIN models also exhibited layer-dependent vulnerability. The findings demonstrate that middle layers were especially susceptible. The last layer showed reduced responsiveness to bit flips, suggesting that earlier transformations carry more influence over final predictions in GIN. 
\par GCN and GAT models demonstrate more moderate vulnerability in compare to GIN and GraphSAGE. Moreover, both models are vulnerable on the deeper layer. GAT, despite its attention-based mechanism, does not exhibit stronger resistance to GBFA than GCN. 
\par \textbf{GBFA Effectiveness Across Datasets:} As Cora and PubMed datasets differ in size, feature sparsity, and graph topology, the results show that GBFA consistently induces more severe degradation on the Cora dataset compared to PubMed, even under the same BER and model configurations. On Cora, a sparse dataset, GBFA is able to significantly reduce model accuracy by flipping a minimal number of bits. Conversely, the PubMed dataset, which features a denser graph and a lower label rate, appears more resilient. Weight distribution analysis further supports this observation.  As illustrated in Figure \ref{4}, models trained on PubMed exhibit broader weight distributions, suggesting higher tolerance to minor perturbations. In contrast, weights on Cora are more tightly clustered, potentially amplifying the effect of small bit-level changes.
%\par Our findings demonstrate the efficacy of the GBFA attack across various GNN models on the Cora and PubMed datasets. Notably, these models exhibit greater susceptibility to the GBFA attack on the Cora dataset. Figure \ref{3} presents the weight distributions of GCN and GAT trained on the Cora and PubMed datasets. The weight distributions for GCN models exhibit a wider spread which can be infered greater flexibility in weight updates, potentially capturing more variance in node feature transformations. In contrast, the weight distribution for GAT models demonstrate a significantly sharper peak around zero, indicating a more concentrated range of weight values. Furthermore, the dataset characteristics influence weight distributions. The Cora dataset leads to a more constrained weight range compared to PubMed, which exhibits a wider distribution in both GCN and GAT models.  Additionally, GIN and GraphSAGE exhibit greater vulnerability to the GBFA attack, with the PAC reaching 13\%.
\begin{table}[h]
    \centering
    \caption{Post-Attack Test Accuracy of Random Bit-Flip Fault Attack and GBFA on GNNs using Cora and PubMed datasets at the Minimum BER}
    \renewcommand{\arraystretch}{1.2}
    \setlength{\tabcolsep}{2pt} % Adjust column spacing for better fit
    \scriptsize % Reduce font size for fitting in one column
    \begin{tabular}{|c|c|c|c|}
        \hline
        \textbf{Model} & \textbf{Random} & \textbf{GBFA} & \textbf{Min} \\
         & \textbf{Bit-Flip} & \textbf{Attack} & \textbf{BER} \\
        \hline
        GAT-Cora & 74\% & 51\% & 1e-4 \\
         \hline
        GCN-Cora & 81\% & 80\% & 1e-3  \\
        \hline
        GCN-PubMed & 78\% & 77\% & 1e-4 \\
        \hline
        GraphSAGE-PubMed & 76\% & 75\% & 1e-3 \\
        \hline
        GIN-Cora & 64\% & 63\% & 1e-2 \\
        \hline
        GIN-PubMed & 61\% & 60\% & 1e-2 \\
        \hline
    \end{tabular}
    \label{T9}
\end{table}

\subsection{Comparison with other methods}
In this section, we compare our proposed GBFA attack with random bit flip attack and recent existing work of attack.
\par In the random bit-flip attack, the attacker randomly selects weights in the network without considering layer awareness, using the minimum BER that causes a drop in accuracy. Table \ref{T9} presents the post-attack test accuracy for both the random bit-flip attack and GBFA at the minimum BER. The results demonstrate that GBFA consistently achieves slightly lower accuracy than random bit-flip attacks, suggesting its effectiveness in targeting model vulnerabilities. Overall, GBFA slightly outperforms random attacks in degrading accuracy, with the largest drop observed for GAT-Cora (51\% vs. 74\%).
%The IBFA attack in \cite{kummer2024attacking} is similar to our proposed GBFA attack, as both employ bit-flip faults to compromise GNNs.
IBFA introduced in \cite{kummer2024attacking} is another attack on GIN model. It introduces a theoretically motivated bit-flip attack specifically tailored to degrade the expressivity of GIN by targeting the injectivity of aggregation function, our proposed GBFA attack focuses on flipping trained weight parameters.  Unlike IBFA, which assumes full white-box access and leverages properties of the Weisfeiler-Leman (1-WL) test to disrupt graph isomorphism distinctions, GBFA operates under a gray-box assumption, using runtime memory access patterns to infer layer execution and identify vulnerable bits via gradient-guided in-layer search. IBFA is applied only GIN model while our proposed GBFA attack can be applied on different GNN models. On the other hand, as they applied different datasets it is impossible to compare GBFA attack on GIN and IBFA on GIN.

% \section{Discussion}
%The proposed GFFA attack operates without hardware support and therefore incurs no area overhead. \textcolor{red}{I don't think so. If you do not have hardware support, how will you flip the bits? } 
%\textcolor{blue}{I removed hardware overhead of the GBFA}

\par \textbf{Time complexity analysis of the GBFA attack.} For each iteration of GBFA to identify single most vulnerable bit in a selected weight in the targeted layer, the time complexity is $\mathcal{O}(N)$, where N is the number of bit with highest gradient ranking that will be detected in GBFA method. In general, the time complexity of the proposed GBFA is linear for each iteration.
%For each iteration, the attacker involves selecting a subset of weights to modify, which requires linear time with respect to the specified BER. We assume that $n$ is the number of weights targeted to attack in targeted layer, so time complexity is $\mathcal{O}(W)$  
 %In general, the time complexity of the proposed GBFA is linear for each iteration.
% For each iteration of GBFA to identify single most vulnerable bit in in-layer search, the time complexity is $\mathcal{O}(N)$. $N$ is the number of bits with highest gradient ranking that will be checked in GBFA method.
\section{Conclusion}
%\section*{References}
In this paper, we present GBFA, a hardware-based fault injection attack targeting GNNs deployed on the hardware accelerator for node classification tasks. The attack degrades prediction accuracy by selectively flipping bits in the model’s stored weights. GBFA leverages memory traces and a Markov model to predict the execution order of layers. In the identified target layer, it flips a vulnerable bit within a selected weight using a gradient descent-based ranking method. Weights are gradually selected based on the BER, ensuring that model accuracy degrades with minimal bit perturbation. We evaluate the effectiveness of  GBFA  on four GNN architectures: GCN, GAT, GraphSAGE, and GIN, using the Cora and PubMed datasets. Our results demonstrate that GBFA can successfully compromise all the GNN models. Furthermore, comparing the GBFA results with those of a method that injects bit-flip faults without targeting a specific layer highlights the significance of in-layer bit-flip fault injection attacks in compromising model accuracy. Our research highlights the importance of addressing hardware-based attacks and underscores the need for dedicated defense mechanisms to ensure the reliable deployment of GNN models in safety-critical and real-world applications.

% \textcolor{red}{Please do few changes: 1. Add 4-5 references from our group. 2. Change title a bit and/or abstract. You can use LLMs to do this rephrasing.}
% \textcolor{blue}{I cited four works from our group, rewrote the entire abstract, changed the title, and updated Table VI to include the GraphSAGE model.}
\bibliographystyle{acm}
\bibliography{paper01}
\end{document}